%% file: main.tex
\documentclass[sigconf]{acmart}  
\usepackage[utf8]{inputenc}
\usepackage{setspace}
\usepackage{kotex}
\usepackage{graphicx}
\usepackage{subcaption}
\usepackage{amsmath, amsthm} 
\usepackage{tikz}
\usepackage{lipsum}
\usepackage{booktabs}
\usepackage{enumitem}
\usepackage{multirow}
\usepackage{makecell}
\setlist{nolistsep}
\AtBeginDocument{%
  }

\setcopyright{none}

\acmConference[KDD'23 Workshop on Machine Learning in Finance
]{}{Aug 06--08, 2023}{Long Beach, CA}

\begin{document}
\newcommand{\cy}[1]{\textcolor{violet}{#1}}
\newcommand{\sd}[1]{\textcolor{magenta}{#1}}
\setlength{\abovedisplayskip}{3pt}
\setlength{\belowdisplayskip}{3pt}

\title{Customs Import Declaration Datasets}

\author{Chaeyoon Jeong}
\affiliation{%
  \institution{KAIST}
  \city{Daejeon}
  \country{South Korea}}
\email{lily9991@kaist.ac.kr}

\author{Sundong Kim}
\affiliation{%
  \institution{GIST}
  \city{Gwangju}
  \country{South Korea}}
\email{sundong@gist.ac.kr}
\authornote{Corresponding author}

\author{Jaewoo Park}
\affiliation{%
  \institution{Korea Customs Service}
  \city{Daejeon}
  \country{South Korea}} 
\email{jaeus@korea.kr}

\author{Yeonsoo Choi}
\affiliation{%
  \institution{Korea Customs Service}
  \city{Daejeon}
  \country{South Korea}} 
\email{yschoi0817@gmail.com}

\renewcommand{\shortauthors}{Jeong et al.}

\begin{abstract}
Given the huge volume of cross-border flows, effective and efficient control of trade becomes more crucial in protecting people and society from illicit trade. However, limited accessibility of the transaction-level trade datasets hinders the progress of open research, and lots of customs administrations have not benefited from the recent progress in data-based risk management. In this paper, we introduce an import declaration dataset to facilitate the collaboration between domain experts in customs administrations and researchers from diverse domains, such as data science and machine learning. The dataset contains 54,000 artificially generated trades with 22 key attributes, and it is synthesized with conditional tabular GAN while maintaining correlated features. 
Synthetic data has several advantages. First, releasing the dataset is free from restrictions that do not allow disclosing the original import data. The fabrication step minimizes the possible identity risk which may exist in trade statistics. Second, the published data follow a similar distribution to the source data so that it can be used in various downstream tasks. Hence, our dataset can be used as a benchmark for testing the performance of any classification algorithm. With the provision of data and its generation process, we open baseline codes for fraud detection tasks, as we empirically show that more advanced algorithms can better detect fraud.

\end{abstract}



\maketitle

\begin{figure}[t]
\centering
    \includegraphics[width=0.8\linewidth]{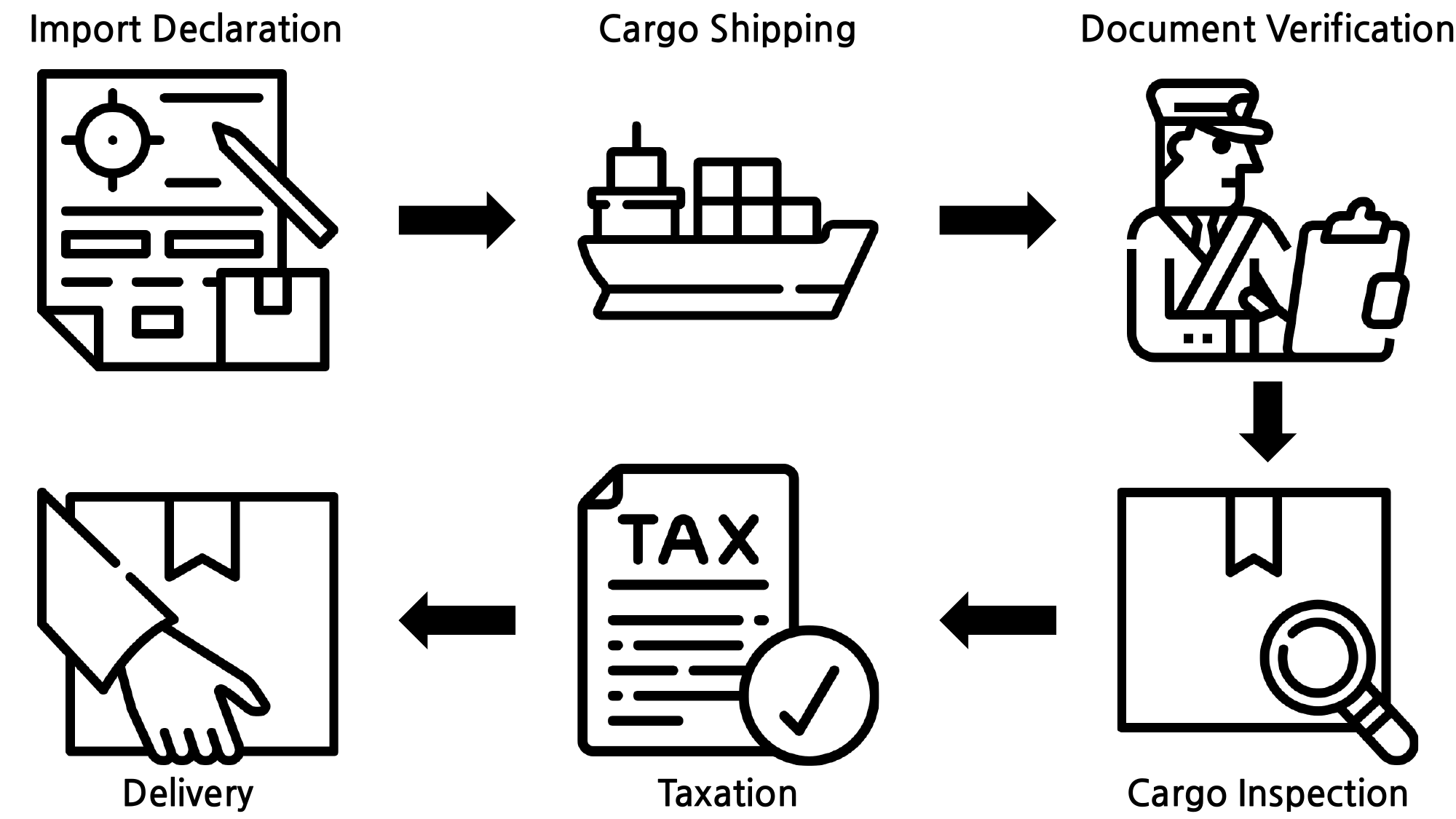}
    \caption{Import clearance process}
     \label{fig:process}
     \vspace{-3mm}
\end{figure}

\section{Introduction}
Customs clearance is the process of getting permission from customs administrations to either move goods out of a country (export) or bring goods into the country (import). The customs declarant declares the goods to the customs office, and permission is given only when the declaration is legitimate. In the case of imports, if the value of the shipment exceeds the threshold (\$150 in Korea), the customs impose tariffs on the item. Once the tariff is collected, the goods are allowed to be released. 

Despite the enthusiasm around the use of data and the possibilities offered by artificial intelligence~\cite{mikuriya2020wco}, the adoption of new technology is relatively slow in the customs community. The primary reason is the lack of publicly available data. Disclosure of import declaration data outside customs is strictly prohibited because of its confidentiality.
Only authorized departments or institutions could conduct research internally, and there is no visible community effort. 

To address this challenge, we are inspired by the recent use of data generation techniques in other domains, such as medical information~\cite{chen2021synthetic}. Since such synthetic datasets have similar distributions to raw datasets, those can be used to train machine learning models and perform various data-driven analysis tasks. This approach leads us to design synthetic data that can be open to the public. 

\begin{figure*}[t]
\centering
    \begin{subfigure}[t]{.44\linewidth}
        \centering
        \includegraphics[width=\linewidth]{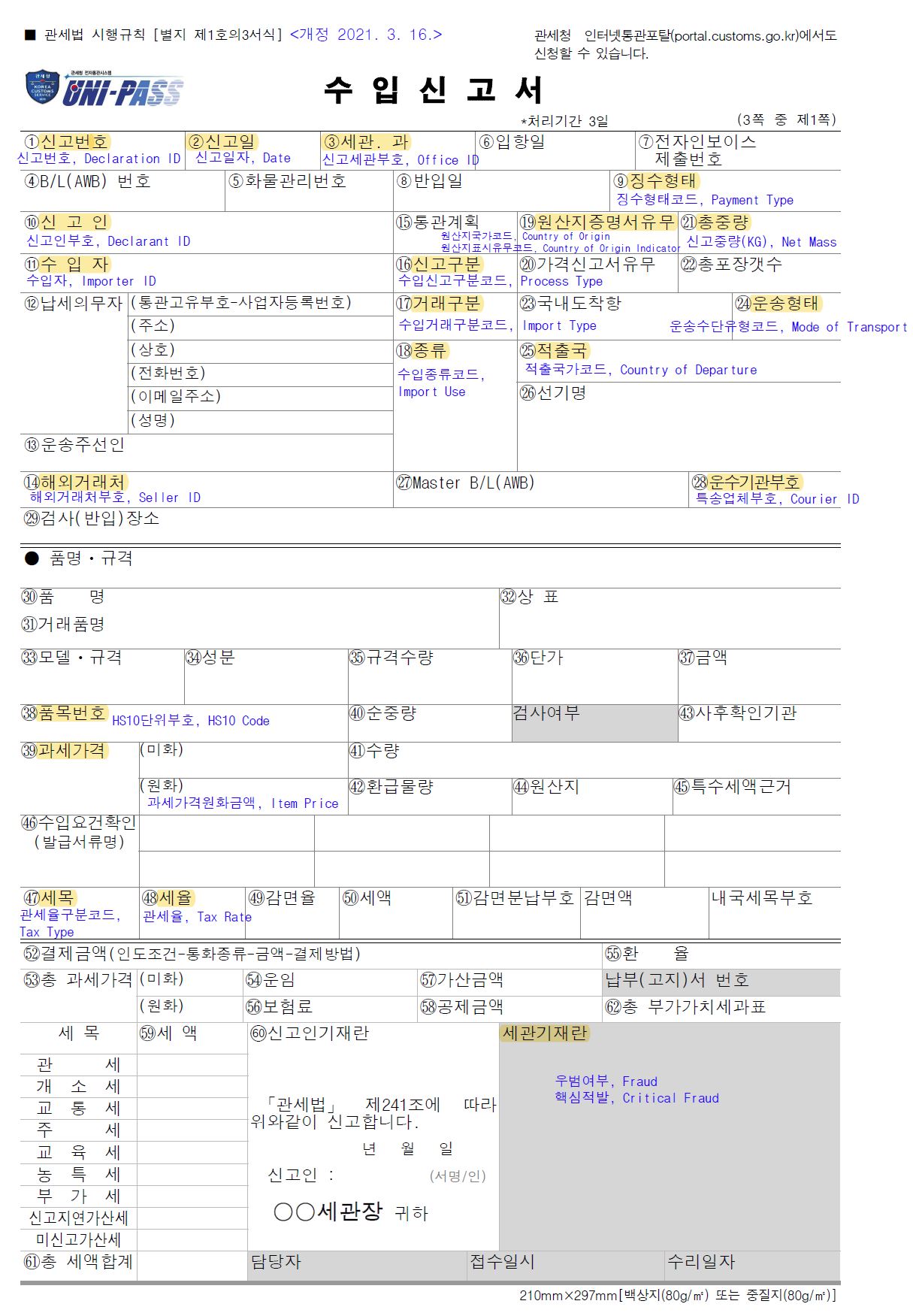}
        \caption{Import declaration form (Korean)}
    \end{subfigure}
    \begin{subfigure}[t]{.44\linewidth}
        \centering
        \includegraphics[width=\linewidth]{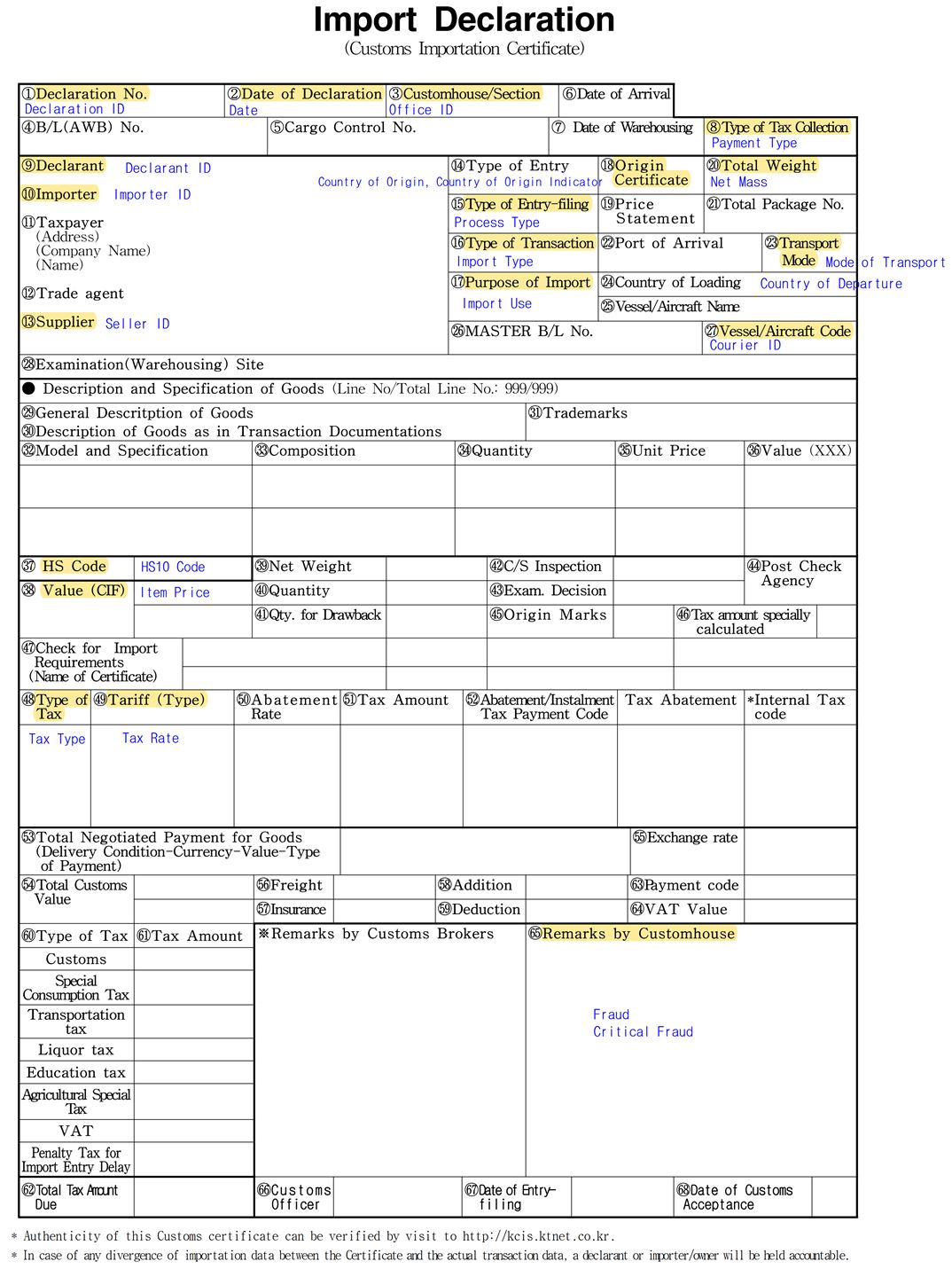}
        \caption{Import declaration form (English)}
    \end{subfigure}
    \caption{The customs clearance processes, including as customs declaration, tax payment, and application requirements needed by people and corporations when exporting or importing, are handled via the online system UNI-PASS. It enables the KCS to electronically process 430 million declarations and 50 million travelers per year. In the above import declaration form, we picked key attributes and synthesized data as in Table~\ref{tab:datadescription}.}
    \label{fig:form}
\end{figure*}

The dataset introduced in this paper includes 54,000 artificially generated trades with 22 attributes. The 22 columns refer to each entry in an import declaration form. Using a tabular synthesizer with post-processing techniques, we maintain that the distribution and correlation among features in the synthetic dataset are similar to those of the source dataset. Refer to Sections~\ref{sec:data_description} and \ref{sec:data_generation} for more information.

Additionally, our dataset is being used as a valuable resource for applied AI researchers, including those in the field of customs, government agencies, or finance industries. The dataset mitigates privacy concerns and enables more accurate and fair comparisons of different fraud detection algorithms. Our synthetic dataset can help researchers who require access to realistic data since real-world customs data cannot be released due to privacy concerns. Also, it can be used for benchmarking and testing machine learning algorithms. Any algorithm can be evaluated by fraud detection problem or HScode classification using our tabular dataset. As shown in Section \ref{sec:fraud_detection}, more advanced algorithms tend to achieve better precision than relatively conventional models. This implies the dataset can serve as a useful benchmark for evaluating the performance of various algorithms. For more details, refer to Section~\ref{sec:fraud_detection}.

Moreover, customs agencies can develop their capabilities in data science through accessible synthetic data and data generation techniques. For instance, the World Customs Organization (WCO) is using synthetic data for educational purposes. In their Advanced Data Analytics course, synthetic customs declaration data is utilized for training fraud detection models such as LITE DATE\footnote{An analytic model for fraud detection and advanced online course from WCO. \\ \url{https://bit.ly/3LtJ8Qg}} or DATE~\cite{DATE}. Also, the data and its synthesizing step have been used as learning material for the WCO members, three-quarters of which are developing countries.\footnote{WCO PICARD 2022 hands-on workshop. \url{https://bit.ly/3njmYH0}} Meanwhile, the data is used for competition between universities to prototype fraud-detection algorithms that can be applied in practice to detect illicit imports.\footnote{Competition held at CNU. \url{https://bit.ly/KCS-CNU-Competition}} Consequently, development in data analysis skills can facilitate the field in making new research questions and hypotheses.

We conclude the paper by discussing possible scenarios to use the data and summarizing necessary thoughts on the data synthesis. 
The data and code can be found in \url{https://bit.ly/customs-dataset}. 

\input{tables/data_description}

\section{Related Works}
\textbf{Data Synthesis}: While artificial intelligence (AI) is bringing us remarkable achievements in numerous domains, a high-quality dataset is crucial for developing a good AI model. However, there are many obstacles in utilizing a raw dataset, such as privacy concerns, data paucity, and data bias \cite{gianfrancesco2018potential, mehrabi2021survey}. Accordingly, generative methods are in the limelight as a method to solve these problems \cite{nikolenko2021synthetic, yoon2020anonymization, jordon2019pate,yoon2022ehr, qian2023synthcity, chen2021synthetic,hwang2022}. Note that the synthesized output with the generative model has a similar distribution as the input data, though the generated data is not real data. Therefore data synthesis approaches can provide not only a large number of high-quality data but also are able to generate datasets with improved responsibility, fairness, privacy, and robustness, leading AI models to inherit these properties during training.
\smallskip

\noindent\textbf{Generative Adversarial Network}: A typical data generation method is deep-learning-based generation method. Especially, Generative Adversarial Networks (GAN) are widely used for synthesizing various formats of data \cite{goodfellow2020generative, kong2020hifi, yoon2019time,de2019training}. Due to its adversarial architecture, It has the ability to learn the pattern and distribution of original data with additional restrictions or conditions. For example, many works adapt GAN to generate synthetic data in order to address privacy or robustness concerns in medical or health data \cite{yoon2022ehr,islam2020gan,chen2021synthetic}. Tabular data is multi-modal data: each attribute (column) has different properties, which are unique characteristics distinct from other data modalities (e.g., image, text). Some columns are continuous while others have discrete values. Variations such as conditional tabular GAN (CTGAN), variational autoencoder (VAE) for mixed-type tabular data generation (TVAE), and Tabular GAN (TGAN) are specialized for this data format, outperforming conventional data generation techniques \cite{xu2018synthesizing, CTGAN}.

\section{Data Description}
\label{sec:data_description}
In this section, we illustrate the layout and characteristics of the data, including definitions of customs-specific terminology. Additionally, we demonstrate the similarities between original source data and synthetic data. This shows that synthetic data can serve as a good alternative to real-world Customs declaration data.

\subsection{Data Schema}
The tabular form dataset consists of 54,000 import declarations, where each row describes the report of a single item. 
Among 62 attributes in the import declaration form,\footnote{Import declaration format is shown in Figure 2. More explanation is available at \url{https://bit.ly/import-declaration-form}.} The data includes 22 representative attributes without overlapped or less essential ones. The first 20 values are filled in by importers at the declaration stage of customs clearance, while the rest two attributes are labeled after the customs inspection. 
Categorical attributes and their values follow the handbook provided by the Korea Customs Service (KCS), which contains trade codes used for filling out import and export declarations in Korea.\footnote{The handbook is available at \url{https://www.data.go.kr/data/3040477/fileData.do}.} 
Detailed data descriptions and example values are shown in Table~\ref{tab:datadescription}. 

\emph{Fraud} indicates whether the inspected result of the actual imported goods conflicts with its declaration. \emph{Critical fraud} is a case that may threaten society's public safety or stability, such as copyright infringement, tax evasion, drug smuggling, or false declaration of the origin of goods. In detail, KCS operates a risk management system to detect suspicious imported goods. This system uses either computer-based or human-sampling methods to identify items for inspection. Computer-based sampling uses pattern analysis or machine learning-based algorithms to identify potentially illegal cargo, while manual selection is done by customs officers working on-site. Selected cargos undergo on-site inspection, and the inspection results are denoted using a standardized code system. The result codes indicate the type of violation, including improper classification of goods, false price declaration, quantity discrepancies, incorrect application of tariff rates, and false country of origin labeling. Depending on the severity of each inspection result, the cargo is labeled as either \emph{Normal}, \emph{Fraud}, or \emph{Critical Fraud}.

\input{figure-texts/eda4}

\subsection{Data Reliability}
Statistical test results indicate that the synthetic data and the source data are drawn from similar distributions. The quality of the synthetic tabular data was evaluated using metrics provided by the SDMetrics library from the Synthetic Data Vault project \cite{sdmetrics}.\footnote{For more detail, please refer to \url{https://docs.sdv.dev/sdmetrics/}} The summary of the evaluation results are presented in Table \ref{tab:quality_metrics}. Note that all statistical tests return scores ranging from 0.0 to 1.0, where a higher value means that the synthetic data has better quality in terms of column distribution, the relationship between attributes, the validity of numbers, and diversity. \smallskip

\input{tables/quality_metrics}
\noindent\textbf{Column Shape Similarity}: The similarity between the distribution of each corresponding column in the original and synthetic data was assessed using the Kolmogorov-Smirnov statistic for numerical columns and the total variation distance for categorical columns. In our data, we regard \emph{Date, Tax Rate, Net Mass, Item Price} as numerical columns, and all other attributes as categorical columns.

The Kolmogorov-Smirnov test transforms each numerical column $C$ into a cumulative distribution function (CDF), calculates the maximum distance $\delta$ between the CDFs of the real ($r$) and synthetic ($s$) data, and scales this to a 0-1 range. Column similarity is then derived as $1-\delta$. The final similarity score averages this across all numerical columns, which was 0.8268.

For categorical column $C$, the total variation distance $\delta$ measures the difference between the normalized frequencies $f(x)$ of each value $x$. The similarity score between the real data $r$ and the synthetic data $s$ can be calculated as:
\begin{equation}
\label{eq:tvd}
    \text{score}(C) = 1 - \delta(f_r - f_s) = 1 - \frac{1}{2}\sum_{x \in C} | f_r(x) - f_s(x) |,
\end{equation}
where $x$ represents unique value in column $C$, $f_r(x)$ and $f_s(x)$ denote to normalized frequencies of categorical value $x$ in the real data $r$ and synthetic data $s$, respectively. The average similarity score across all categorical columns was 0.8919.

Figure~\ref{fig:eda1} shows the similarity of column distribution of representative features (\emph{Tax Rate, Net Mass, Critical Fraud}). \smallskip  
\input{figure-texts/eda1}

\noindent\textbf{Column Pair Trend Similarity}: To evaluate if synthetic data preserve attribute associations, we analyzed trends between column pairs. Different metrics were used based on the column types. Pearson correlation coefficient measured correlation for numerical pairs, while a contingency metric assessed relationships between categorical columns or a numerical-categorical pair. Specifically, for any numerical column pair $C$ and $C^\prime$, Pearson correlation coefficients were computed for both source and synthetic datasets, leading to the similarity score:
\begin{equation}
\text{score}(C, C^\prime) = 1 - \frac{|\rho_{C_r, C^\prime_r} - \rho_{C_s, C^\prime_s}|}{2}
\end{equation}
Here, $\rho_{C_r, C^\prime_r}$ and $\rho_{C_s, C^\prime_s}$ denote Pearson correlation between columns $C$ and $C^\prime$ in source $r$ and synthetic $s$ data, respectively. The final score is the average across all numerical pairs, yielding 0.9569.

The contingency similarity metric assesses the resemblance between contingency tables of any categorical column pair in the original and synthetic data. If a column is numerical, it is discretized into bins to convert to a categorical format. For columns $C$ and $C^\prime$, we first calculate the normalized proportion of data points for each category value combination in $C$ and $C^\prime$. Total variation distance (Equation \ref{eq:tvd}) then determines the similarity score as:
\begin{equation}
\text{score}(C, C^\prime) = 1- \frac{1}{2}\sum_{x \in C} \sum_{y \in C^\prime} | f_r(x, y) - f_s(x, y) |
\end{equation}
Here, $x$ and $y$ are all possible values in $C$ and $C^\prime$. $f_r(x, y)$ and $f_s(x, y)$ represent joint proportions for categories $x$ and $y$ in the source $r$ and synthetic $s$ data, respectively. The average contingency table similarity was found to be 0.7633.

The similarity scores across all column pairs are displayed in Figure~\ref{fig:column_pair_trend}. This visualization highlights that the inter-column relationships in the two datasets are largely well-preserved, with the exception of anonymized columns like \emph{Seller ID} or \emph{Importer ID}. The impact of such anonymization is further discussed in Section~\ref{sec:preprocessing_anonymization}.\smallskip
\input{figure-texts/column_trends}

\noindent\textbf{Data Coverage}: This analysis investigates the extent to which synthetic data can reflect the diverse values found in real data. It applies two distinct measures, one for numerical columns and the other for categorical variables. Each measure produces a score where 1.0 indicates perfect value coverage, meaning every value present in the original data is also found in the synthetic data. 

For numerical columns, the range coverage test inspects how the synthetic column's minimum and maximum values align with those of the respective real column. The score for column $C$ is computed as:
\begin{equation}
\begin{split}
\text{score}(C)= 1 - \left[ \text{max}\left(\frac{\text{min}(C_s) - \text{min}(C_r)}{\text{max}(C_r) - \text{min}(C_r)}, 0\right) \right. \\
\left.+ \text{max}\left(\frac{\text{max}(C_r) - \text{max}(C_s)}{\text{max}(C_r) - \text{min}(C_r)}, 0\right)\right]
\end{split}
\end{equation}
Here, $C_r$ and $C_s$ denote the sets of values in the real and synthetic columns, respectively. It should be highlighted that this metric does not account for whether synthetic values exceed the real data's range. If the synthetic minima and maxima extend beyond those of the real data, this implies full range coverage, yielding a score of 1. Our coverage in this respect was found to be 0.8022.

For categorical variables, the category coverage metric is employed. This method determines the proportion of unique values in the real column $C$ that also appear in the synthetic data. We found our coverage to be 0.8801 in this case. \smallskip

\noindent\textbf{Data Boundary}: The boundary property assesses whether the synthetic data preserves the numerical boundaries observed in the real data while excluding outliers. It calculates the proportion of synthetic numerical values that fall within the range defined by the minimum and maximum values of the corresponding real column. \smallskip

\noindent\textbf{Diversity of Generated Data}: This evaluation verifies the uniqueness of each row in the synthetic data, $s$, by checking if it is a mere duplicate of a row from the source data, $r$. To classify as a duplicate, all the values in a synthetic row, denoted as $s_i$, must exactly match a row in the real data.

The criteria for matching differ for numerical and categorical columns. For categorical data, an exact match between synthesized and real values is required. In contrast, numerical columns are first min-max scaled, and a value is considered a match if it lies within 1\% of a real value.

The diversity score is then calculated as the complement of the proportion of duplicated synthetic data points to the total number of synthetic rows.

\begin{equation}
\text{score} = 1 - \frac{1}{n}\sum_{i=1}^{n} \mathbb{I}(s_i \in r)
\end{equation}

where $\text{score}$ denotes the diversity score of the synthetic dataset, $\mathbb{I}(s_i \in r)$ is the indicator function that equals 1 if synthetic row $s_i$ exactly matches any real row and 0 otherwise, $n$ represents the total number of synthetic rows. The sum in the numerator is calculated over all rows in the synthetic dataset. Our synthetic data show the perfect score of 1.0, indicating that every synthetic row is unique. This high level of uniqueness reduces the risk of exposing sensitive information about individuals and can prevent re-identification or data linkage attacks.

\section{Data Generation}
\label{sec:data_generation}
This section provides a detailed account of how the data was generated, by going over each step and explaining the purpose along with the technical detail. The process of data generation involves several steps, including data preprocessing, column aggregation, training the CTGAN model, and postprocessing of the generated output. 

\subsection{Preprocessing}
\label{sec:preprocessing_anonymization}
Among 24.7 million customs declarations reported for 18 months between January 1, 2020, and June 30, 2021, we used the inspected (\textit{i.e.}, labeled) part of the declarations to synthesize our data. Inspected items account for a relatively small percentage of the total, but they are more accurate, all validated by customs officers. We designate it as the source data throughout the paper. Identifiable information such as the importer name in the source data is anonymized into \emph{Importer ID}. Unlike pseudonymization, anonymization removes the possibility to retrieve the original data. It may eliminate the relationship between other features and each individual, but it is a necessary process to protect personal information. The price of goods traded between vendors (\textit{i.e.}, \emph{Item Price}) can be problematic when fully disclosed, so we add some Gaussian noise to the average price of each category of item. The initial format of the \emph{HS6 Code} column is the 10-digit \emph{HS10 Code}. While the first six digits of HS10 codes are standard worldwide, the remaining four digits are domestic codes used in the Republic of Korea, allowing for more detailed information than the standard 6-digit HS codes.

\subsection{Generating Data with CTGAN}
We used CTGAN~\cite{CTGAN} from the Synthetic Data Vault library to generate the data. CTGAN is specifically designed for tabular data and uses conditional techniques to handle imbalanced discrete and multi-modal continuous variables.
Compared to other tabular generative models such as TGAN~\cite{xu2018synthesizing} or TVAE~\cite{CTGAN}, CTGAN showed the most realistic output to our dataset, preserving the relationship between columns. The data generation process can be done in a serial or parallel manner. Users with limited resources can split the data in chronological order, train the CTGAN model on each split, synthesize samples from each model, and aggregate the result. For each split, the model is trained for 300 epochs. 

\subsection{Maintaining Correlated Attributes}
Tabular data have correlated attributes. For example, attributes \emph{HS10 Code---Country of Departure---Country of Origin---Tax Rate---Tax Type} are highly correlated based on customs valuation policies. To make the import declaration data more realistic, a synthesizer should maintain the correlated attributes and their values during the generation process. 
Although CTGAN is a tabular-specific generative model, the output does not always reflect clear correlations between attributes. To maintain dependencies, we aggregate correlated attributes into a single column and save it temporarily before running the CTGAN model. After running the model, the value is split to have the original format. Another example is \emph{Item Price}, which is an attribute correlated with the \emph{HS10 Code} and the \emph{Net Mass} of an item. To maintain this relationship, \emph{Item Price} is reconstructed after the data generation step by multiplying \emph{Net Mass} and the unit price of each \emph{HS10 Code}. Finally, due to the data 
sensitivity issue, we removed the last 4 digits in \emph{HS10} to convert it to \emph{HS6}. 

\section{Application--- Fraud Detection}
\label{sec:fraud_detection}
This section introduces how our dataset can be used as a benchmark for the customs fraud detection problem~\cite{vanhoeyveld2020belgian}. Note that the goal of using our data in fraud detection is not to train the model on synthetic data and do the inference directly for customs clearance in practice. Instead, it can serve as a useful tool for indirectly comparing the performance of different fraud detection algorithms. In practice, synthetic data is used when collaborating with external parties such as IT professionals and AI researchers, for designing and discussing fraud detection algorithms. Due to data privacy concerns, it is impossible to share real-world customs declaration data outside customs organizations. Therefore, the evaluation and validation of the model are being conducted using synthesized data. Our dataset provides a viable alternative for benchmarking and comparing different fraud detection algorithms without compromising data privacy.

\subsection{Background}
Smuggling and tax evasion are fatal threats to society and customs administrations prevent those risks through customs control. Due to high trade volume and limited resources (\textit{i.e.} budget, number of officers), it is difficult to conduct an exhaustive inspection of all items, so customs define a set of rules to screen out high-risk items based on the contents of import declarations. Therefore, establishing an intelligent customs selection or fraud detection system is key to facilitating the customs clearance process~\cite{DATE, gATE, mai2021drift, kim2021KCS}. By predicting likely fraudulent items, customs authorities can determine the inspection level of each item, with the most suspicious items requiring physical inspection by human workers. In other words, the smarter the algorithm, the more efficiently Customs can operate its workforce. Accordingly, we define the problem as finding a set of highly-suspicious items which is regarded as the target of human inspection.


\subsection{Using the Data}
The fraud detection problem aims to find the patterns behind the features in predicting the target label \emph{Fraud}. Data is split into three pieces. We assign the first 12 months of data to the training set, the following three months to the validation set, and the last three months to the test set. 
Categorical variables are label-encoded and numerical variables are min-max scaled. We apply various models including Logistic Regression, Decision Tree, Random Forest, AdaBoost in scikit-learn~\cite{scikit-learn} and gradient boosting decision tree (GBDT) models such as LightGBM~\cite{LightGBM}, CatBoost~\cite{liudmila2018nips}, XGBoost~\cite{tianqi2016}. We set the model to predict each record's fraud score ranging from 0 to 1. Among test records, $n$\% of items with the highest fraud score are inspected. Model performance is evaluated by the precision@$n$\%, representing how many inspected items are actually a fraud.\footnote{The amount of workforce available for physical inspection is usually fixed, so it is important to achieve the best performance within the limited inspection capacity, without changing $n$. Therefore, precision@$n$\% is a more suitable metric than AUC or f-score.}

\input{figure-texts/eda3}
\input{figure-texts/shap_plot}

\subsection{Performance Comparison}
Table~\ref{tab:precision} shows the performance trend of applying various fraud detection algorithms on synthetic and source data. The results are averaged over five runs. Given that customs administration inspects a limited quantity of goods, we considered two inspection rates -- 5\% and 10\%. In both datasets, precision on the 5\% setting is higher than that of 10\%, and the performance of GBDT models such as CatBoost, XGBoost, and LightGBM tends to be higher than the other models. Interestingly, the performance gained by applying an advanced model is distinguishable in synthetic data with a low inspection rate setting. We conclude from these results that the synthetic data can be used as an open benchmark to develop advanced fraud detection algorithms.

In addition, we compared representative features in the downstream fraud detection task performed on each dataset by using the XGBoost model~\cite{tianqi2016} as illustrated in Figure~\ref{fig:eda3}. The tendency of feature importance score is analogous in both datasets. \emph{Impoter ID, Item Price, Net Mass, Declarant ID}, and \emph{HS6 Code} were considered important with high scores. This indicates that the relationships among attributes that play an important role in the actual customs downstream task are well represented.

\section{Discussion}
In this section, we discuss the potential impact and possible usage of our data, and how the data could be further improved.\smallskip

\input{figure-texts/survey_plots}
\input{tables/compare_performance}
\noindent\textbf{To the Customs Domain}: Synthesizing the import declaration data can greatly benefit the customs community. We introduced our dataset and its related data science technologies to the customs community by organizing a hands-on workshop session at WCO internal event. This session met with an enthusiastic response. As shown in Figure \ref{fig:survey_plots}, most of the participants found the session useful and were eager to participate in similar workshops. This shows that the customs community is looking forward to applying the data and its synthesizing process to capacity building and bringing collaborations. \smallskip

\noindent \textbf{Area of research}: Besides fraud item detection, this import declaration data can be used for solving numerous data science problems in the customs domain such as HS code classification~\cite{lee2021kaia}, trade pattern analysis between the key players such as importers, declarants, and offices. Solving these tasks can also facilitate the customs clearance process or enable reaching out for new research questions. \smallskip

\noindent \textbf{Distribution of data}: To accommodate user convenience, we assumed the generated declarations are all inspected. In detail, the training data was sampled from the labeled instances among the original data, thus the synthetic data is also fully labeled. However, in real-world scenarios, a significant number of goods are processed without undergoing such inspections, especially in developed countries that have low fraud rates~\cite{singh2023graphfc}. Accordingly, real import declaration data is usually partially labeled. To simulate this scenario where data is only partially labeled, we can employ post-processing techniques to erase labels of a portion of data. \smallskip

\noindent \textbf{Degree of fabrication}: Anonymizing the data is insufficient to mitigate the potential risk of releasing the data. Adversaries may catch the patterns between the key players and disrupt the trade order even if the declarant code, product classification code, and extraction country code are anonymized. Therefore, we release synthesized data instead. By generating synthetic data, we ensure that sensitive information is protected while still providing a useful resource for analysis and research.\smallskip

\noindent \textbf{Generative model}: Recently, a diffusion model has been discussed as an alternative way of generating artificial data in the computer vision domain~\cite{dhariwal2021diffusion}. It is drawing attention with highly realistic results. If the diffusion model suitable for tabular data is developed, it can be used for data generation.

\section{Conclusion}
We present the customs import declaration data produced as part of sharing challenging data science problems in customs administration and facilitate the collaboration between customs and data science communities. With a careful fabrication strategy, the generated data is fairly similar to the actual data and can be used as a benchmark for downstream tasks such as fraud detection.

\begin{acks}
This work was supported by the Korea Customs Service, Institute for Basic Science (IBS-R029-C2), and the IITP grant funded by the Korea government (MSIT) (No.2019-0-01842, Artificial Intelligence Graduate School Program (GIST)).
\end{acks}

\balance
\bibliographystyle{ACM-Reference-Format}
\bibliography{references}


\end{document}

%% file: tables/data_description.tex
\begin{table*}[t!]
\centering
\caption{Data description}




\label{tab:datadescription}
\resizebox{\linewidth}{!}{%
\begin{tabular}{ l | c | l } \toprule
    Attribute & Value & Explanation  \\ \midrule
    Declaration ID & 97061800 & Primary key of the record \\
    Date & 2020-01-01 & Date when the declaration is reported \\
    Office ID & 13 & Customs office that receives the declaration (e.g., Seoul regional customs) \\
    Process Type & B & Type of the declaration process (e.g., Paperless declaration)\\
    Import Type & 11 & Code for import type (e.g., OEM import, E-commerce) \\  
    Import Use & 21 & Code for import use (e.g., Raw materials for domestic consumption, from a bonded factory) \\  
    Payment Type & 11 & Distinguish tariff payment type (e.g., Usance credit payable at sight)\\
    Mode of Transport & 10 & Nine modes of transport (e.g., maritime, rail, air)\\
    Declarant ID & L77JJEG & Person who declares the item\\
    Importer ID & HQ0W7JA & Consumer who imports the item \\
    Seller ID & PBP2MYI & Overseas business partner which supplies goods to Korea \\
    Courier ID & MWIDNS & Delivery service provider (e.g., DHL, FedEx) \\ 
    HS6 Code & 090121 & 6-digit product code (e.g., 090121 = Coffee, Roasted, Not Decaffeinated) \\ 
    Country of Departure & JP & Country from which a shipment has or is scheduled to depart \\ 
    Country of Origin & JP & Country of manufacture, production or design, or where an article or product comes from \\ 
    Country of Origin Indicator & B & Way of indicating the country of origin (e.g., B = Mark on package) \\ 
    Tax Rate & 8.0 & Tax rate of the item (\%)\\ 
    Tax Type & A & Tax types (e.g., FTA Preferential rate) \\ 
    Net Mass & 1262.0 & Mass without any packaging (kg)\\ 
    Item Price & 1437418.0 & Assessed value of an item (KRW)\\ \midrule
    Fraud & 1 & Any fraudulent attempt to reduce the customs duty? After inspection, fraud is recorded as 1 (0/1 Binary) \\
    Critical Fraud & 1 & Among frauds, critical frauds that can threaten public safety, are marked as 2 (0/1/2 Ternary). \\ 
    
    \bottomrule
\end{tabular}
}
\end{table*}





    

%% file: figure-texts/eda4.tex
\begin{figure*}[bth!]
\centering
    \begin{subfigure}[t]{.32\linewidth}
        \centering\captionsetup{width=.95\linewidth}%
        \includegraphics[width=\linewidth]{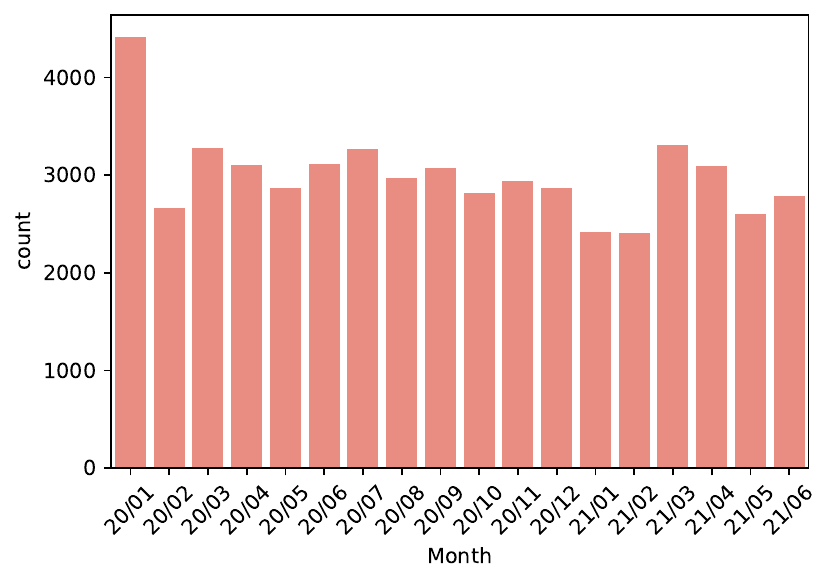}
        \caption{Date}
    \end{subfigure}
    \begin{subfigure}[t]{.30\linewidth}
        \centering\captionsetup{width=.95\linewidth}%
        \includegraphics[width=\linewidth]{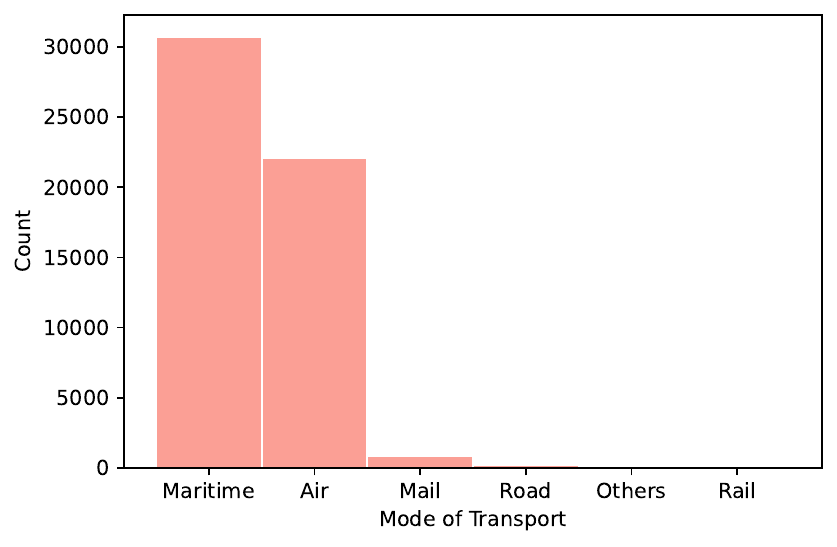}
        \caption{Mode of Transport}
    \end{subfigure}
    \begin{subfigure}[t]{.30\linewidth}
        \centering
        \includegraphics[width=\linewidth]{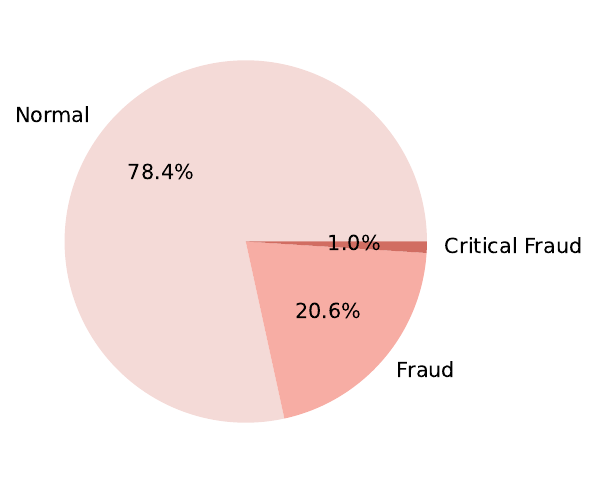}
        \caption{Critical Fraud}
    \end{subfigure}
    \begin{subfigure}[h]{.32\linewidth}
        \centering\captionsetup{width=.95\linewidth}%
        \includegraphics[width=\linewidth]{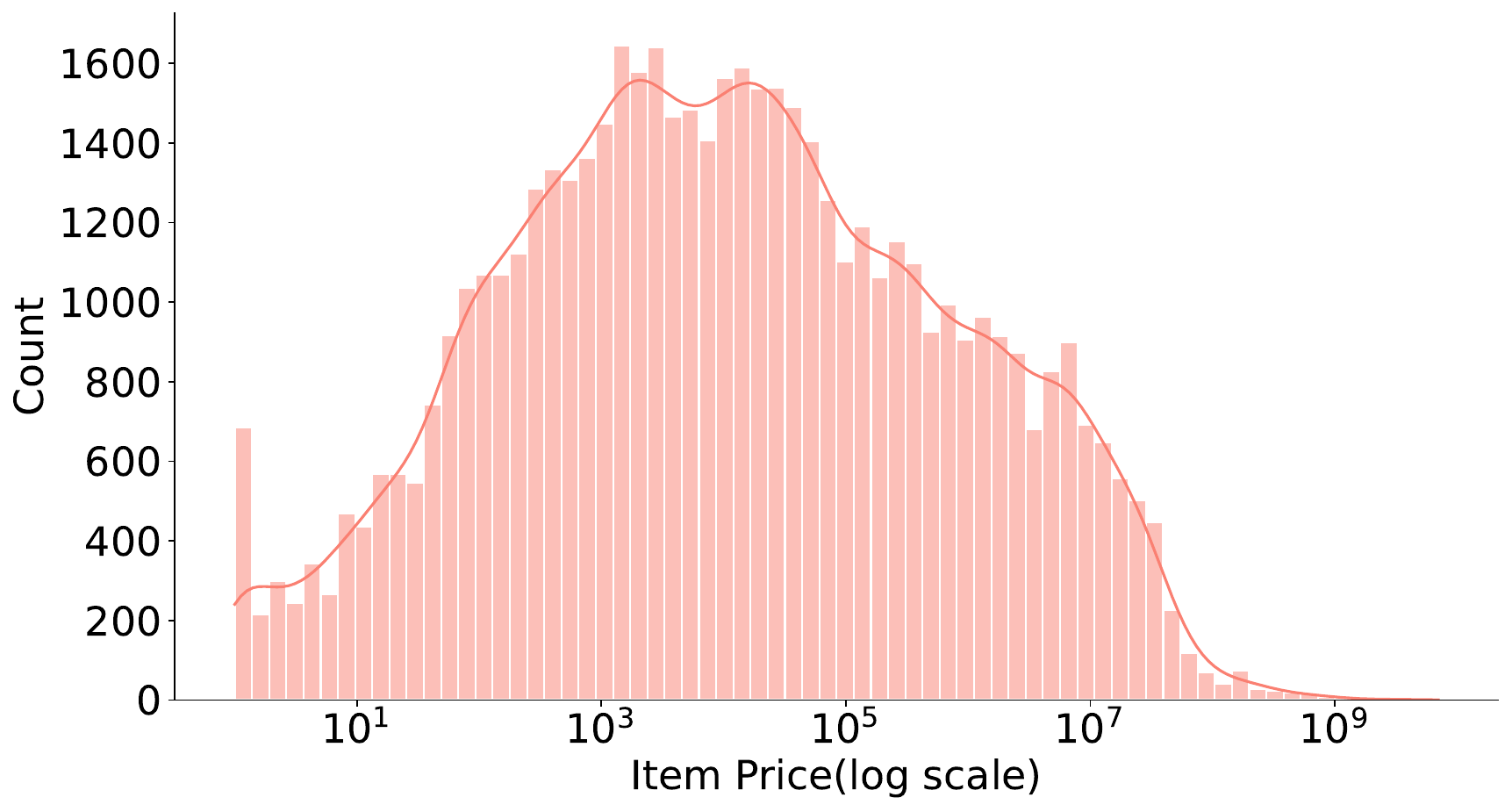}
        \caption{Item Price}
    \end{subfigure}
    \begin{subfigure}[h]{.32\linewidth}
        \centering\captionsetup{width=.95\linewidth}%
        \includegraphics[width=\linewidth]{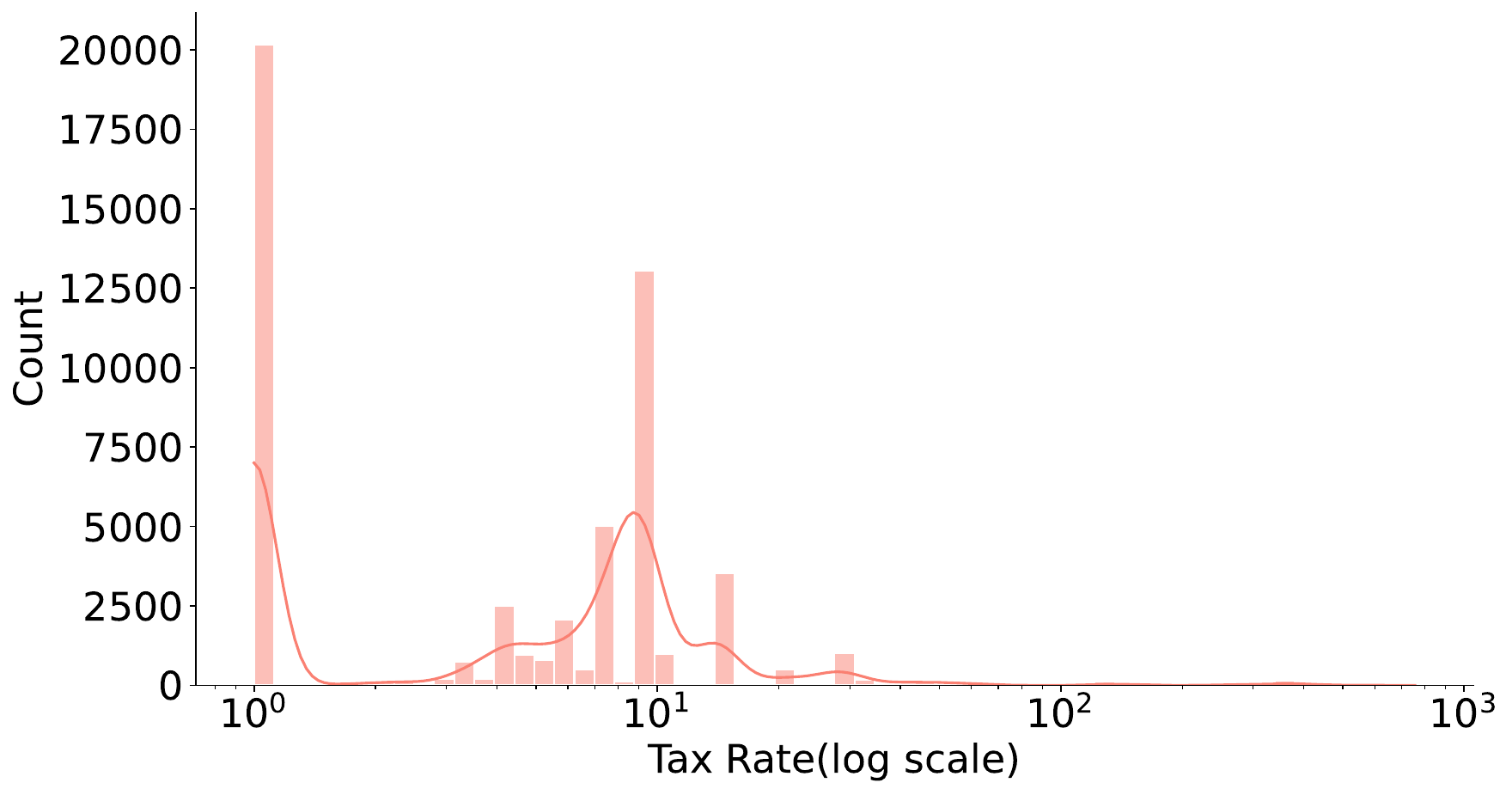}
        \caption{Tax Rate}
    \end{subfigure}
    \begin{subfigure}[h]{.32\linewidth}
        \centering\captionsetup{width=.95\linewidth}%
        \includegraphics[width=\linewidth]{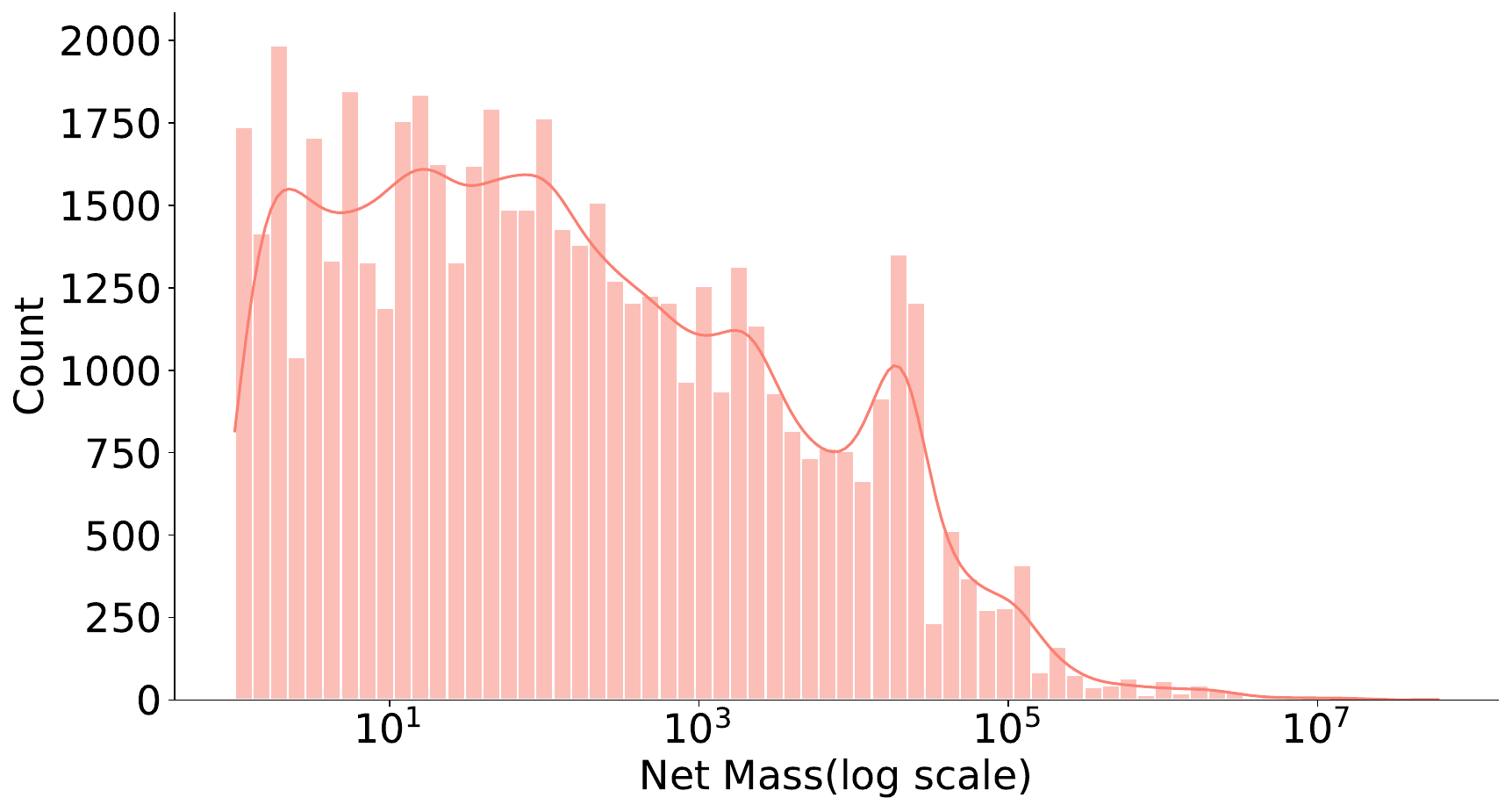}
        \caption{Net Mass}
    \end{subfigure}
    \caption{Exploratory data analysis: Representative attributes and their distribution}
    \label{fig:eda4}
\end{figure*}

%% file: tables/quality_metrics.tex
\begin{table}[]
\centering
\caption{Summarizing synthetic data quality evaluation metrics. The results indicate that the synthetic data and the source data come from similar distributions, while diversity is preserved. Num and Cat refer to numerical and categorical columns, respectively. The score is the average score of all columns corresponding to each type. All scores are between 0.0 and 1.0, and a score close to 1 means that synthetic data shows good quality.}
\label{tab:quality_metrics}
\resizebox{1\linewidth}{!}{
\begin{tabular}{@{}c|c|c|c@{}}
\toprule
\textbf{Property} & \textbf{Metric} & \textbf{Column Type} & \textbf{Score}\\ \midrule
\multirow{2}{*}{Column Shape} & Kolmogorov-Smirnov test & Num & 0.8268 \\ \cmidrule(l){2-4} 
 & Total variation distance & Cat & 0.8919 \\ \midrule
\multirowcell{3.4}{Column Pair Trend} & Person correlation similarity & Num \& Num & 0.9569 \\ \cmidrule(l){2-4} 
 & Contingency table similarity & \begin{tabular}[c]{@{}c@{}}Cat \& Cat or\\ Cat \& Num\end{tabular} &  0.7633 \\ \midrule
\multirowcell{2.4}{Coverage} & Range coverage & Num & 0.8022 \\ \cmidrule(l){2-4} 
 & Category coverage & Cat & 0.8801\\ \midrule
Boundary & Boundary adherence & Num & 0.9869 \\ \midrule
Diversity & New row synthesis & All & 1.0000 \\ \bottomrule
\end{tabular}
}
\end{table}

%% file: figure-texts/eda1.tex
\begin{figure*}[bth!]
\centering
    \begin{subfigure}[b]{.32\linewidth}
        \centering\captionsetup{width=.95\linewidth}%
        \includegraphics[width=\linewidth]{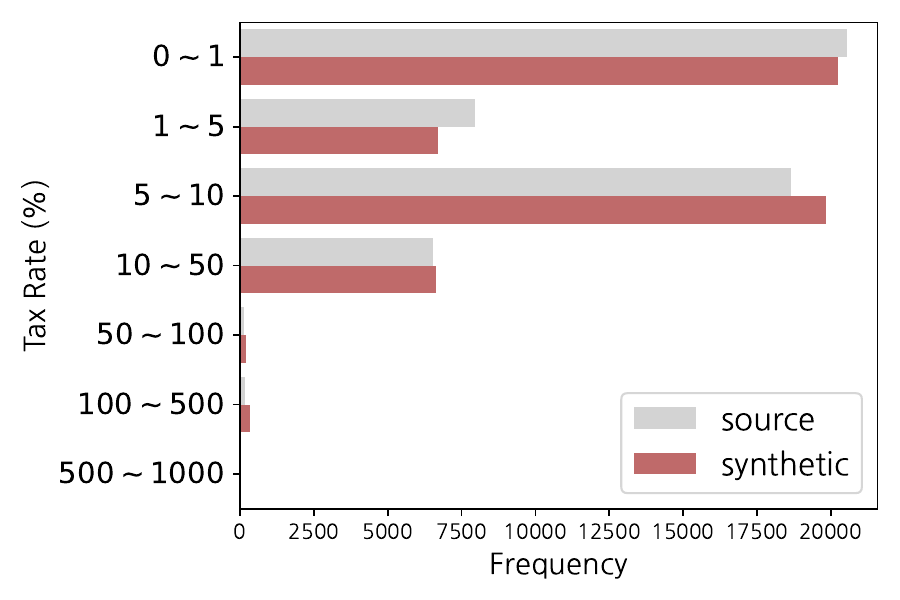}
        \caption{Tax Rate}
    \end{subfigure}
    \begin{subfigure}[b]{.32\linewidth}
        \centering\captionsetup{width=.95\linewidth}%
        \includegraphics[width=\linewidth]{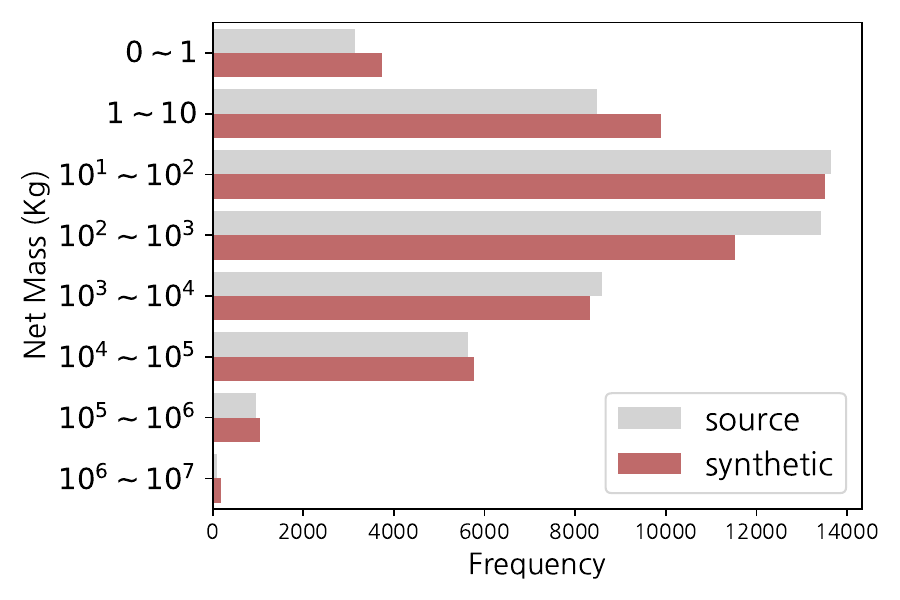}
        \caption{Net Mass}
    \end{subfigure}
    \begin{subfigure}[b]{.32\linewidth}
        \centering\captionsetup{width=.95\linewidth}%
        \includegraphics[width=\linewidth]{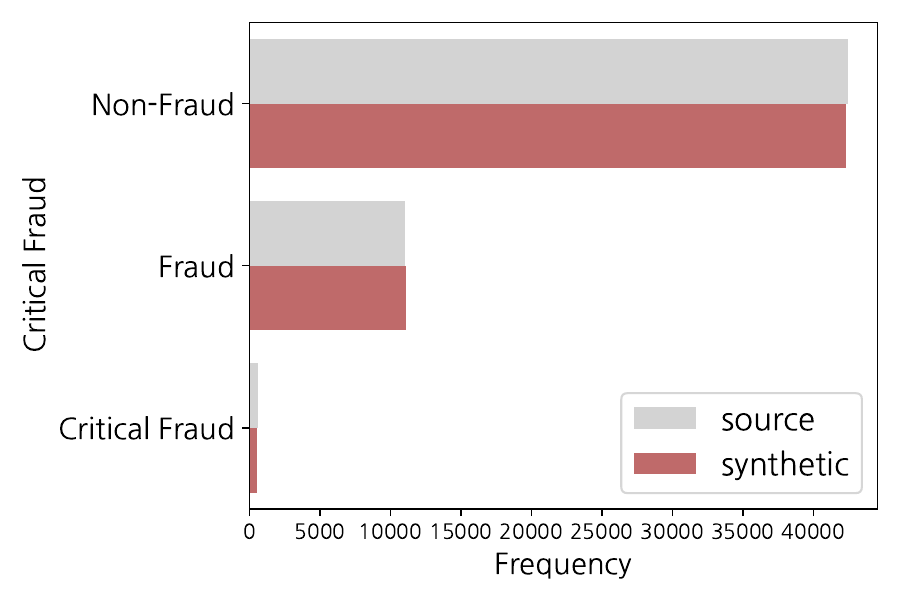}
        \caption{Critical Fraud}
    \end{subfigure}
    \caption{Distribution of representative features is similar between synthetic data and source data.}
    \label{fig:eda1}
\end{figure*}

%% file: figure-texts/column_trends.tex
\begin{figure}[htb!]
\centering
    \includegraphics[width=0.95\linewidth]{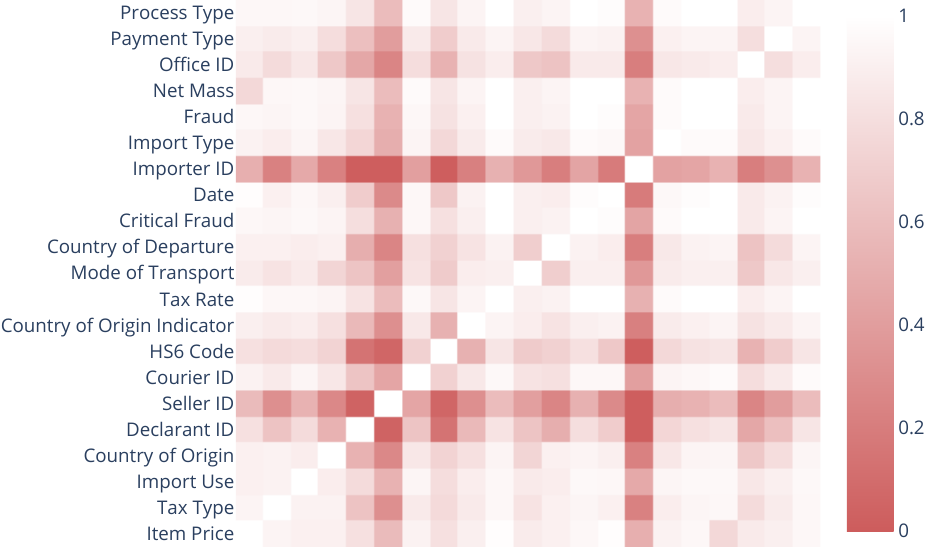}
    \caption{Similarity heatmap of trends between two columns. The white cell indicates the trend between the column pair is similar in real and synthetic data, while the red color indicates they are highly different.}
     \label{fig:column_pair_trend}
\end{figure}

%% file: figure-texts/eda3.tex
\begin{figure*}[htb!]
\centering
    \includegraphics[width=0.7\linewidth]{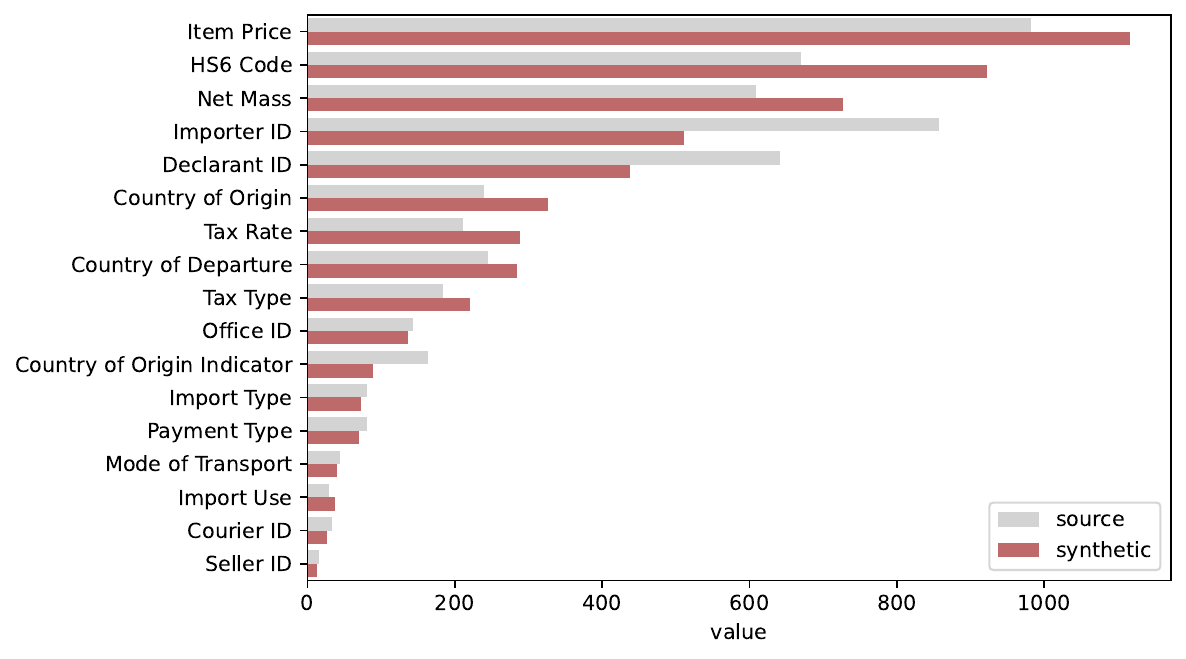}
    \caption{Important features for fraud detection task are also similar between the two datasets (Feature importance score calculated while training XGBoost).}
     \label{fig:eda3}
\end{figure*}

%% file: figure-texts/shap_plot.tex
\begin{figure*}[bth!]
\centering
    \begin{subfigure}[b]{0.49\linewidth}
        \centering\captionsetup{width=.95\linewidth}%
        \includegraphics[width=\linewidth]{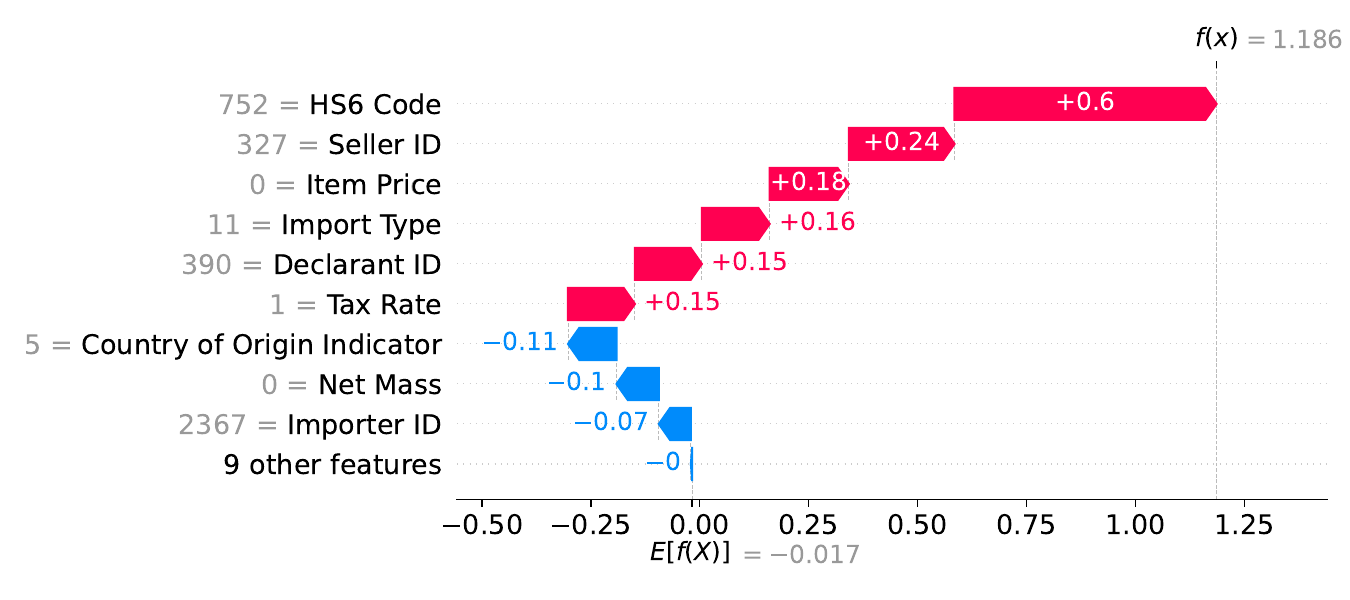}
        \caption{Fraud sample}
    \end{subfigure}
    \begin{subfigure}[b]{0.49\linewidth}
        \centering\captionsetup{width=.95\linewidth}%
        \includegraphics[width=\linewidth]{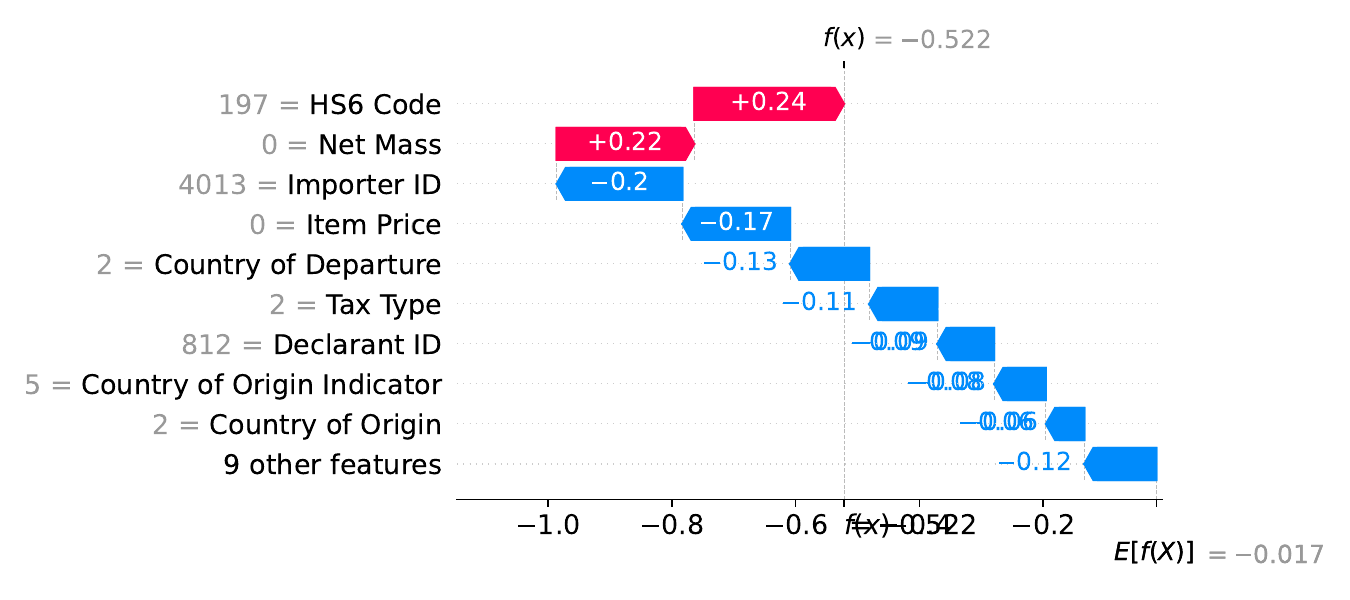}
        \caption{Normal sample}
    \end{subfigure}
    \caption{Shapley values that explain XGBoost decision of a test data instance. The base value $E[f(x)] = -0.017$ refers to the average model output, and each row shows how each feature contributes to the final model output. Red bars indicate that the feature is pushing the prediction higher, while blue bars mean those are forcing the output lower.}
    \label{fig:shap_plot}
\end{figure*}

%% file: figure-texts/survey_plots.tex
\begin{figure*}[t!]
\centering
    \begin{subfigure}[t]{.33\linewidth}
        \centering\captionsetup{width=.95\linewidth}%
        \includegraphics[width=\linewidth]{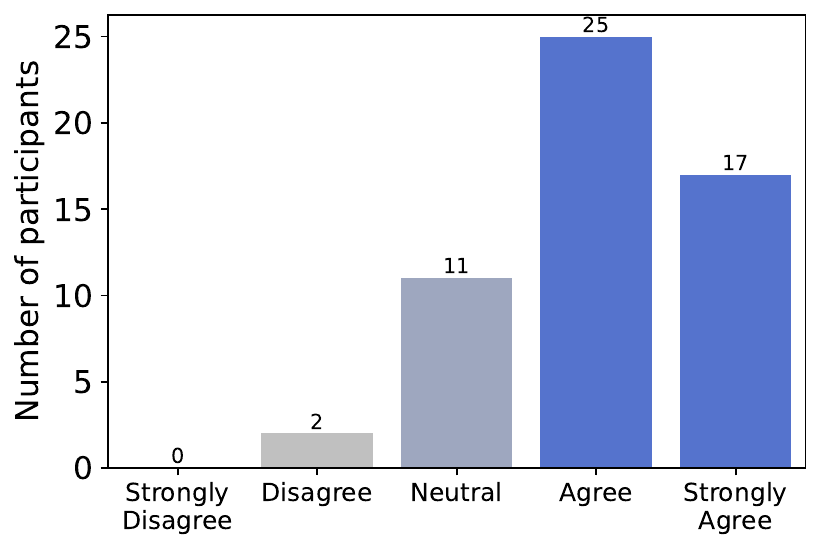}
        \caption{Did the training content meet your expectations and needs?}
    \end{subfigure}
    \begin{subfigure}[t]{.33\linewidth}
        \centering\captionsetup{width=.95\linewidth}%
        \includegraphics[width=\linewidth]{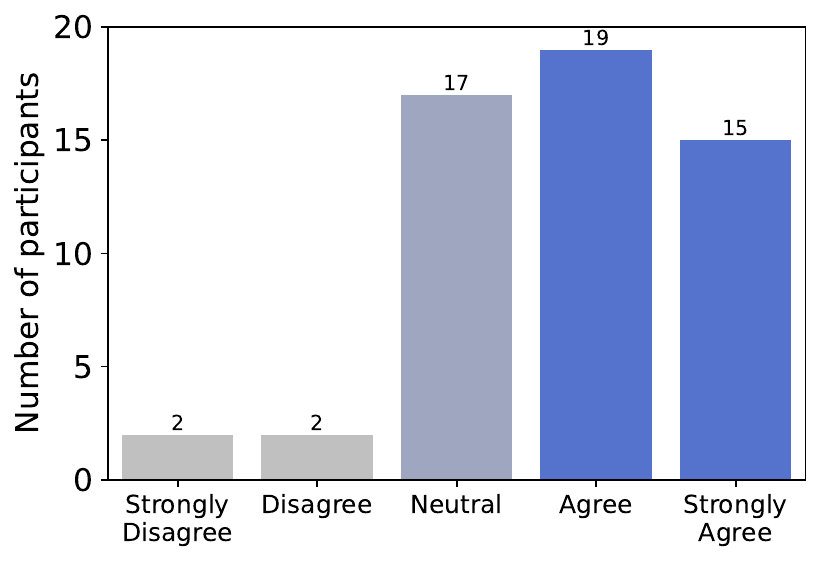}
        \caption{Do you agree the course contents are practical enough to apply in your Customs Organization?}
    \end{subfigure}
    \begin{subfigure}[t]{.33\linewidth}
        \centering\captionsetup{width=.95\linewidth}%
        \includegraphics[width=\linewidth]{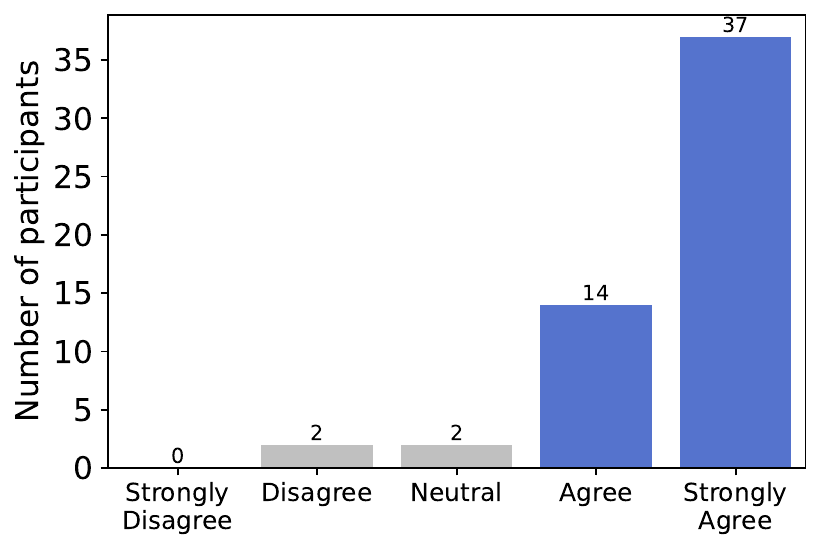}
        \caption{Would you participate again if a similar session takes place next year?}
    \end{subfigure}
    \caption{Survey results of participants' experience in our hands-on data workshop.}
    \label{fig:survey_plots}
\end{figure*}

%% file: tables/compare_performance.tex
\begin{table}[t!]
\centering
\caption{The fraud detection performance in the synthesized data follow a similar trend to the real data.}
\label{tab:precision}
\begin{tabular}{ l | c  c | c  c } \toprule
    \multicolumn{1}{c}{} & 
    \multicolumn{2}{| c |}{Synthetic data} &
    \multicolumn{2}{| c}{Source data}\\ \midrule
    Model | Precision & $n = 5$\% & $n = 10$\% & $n = 5$\% & $n = 10$\%  \\ \midrule
Logistic Regression & 0.2759 & 0.2606 & 0.3921 & 0.3859 \\ 
AdaBoost & 0.3608 & 0.3113 & 0.4902 & 0.4896 \\ 
Decision Tree & 0.3561 & 0.3196 & 0.5128 & 0.4600 \\ 
Random Forest & 0.3608 & 0.3420 & 0.5035 & 0.4739 \\ 
CatBoost & 0.6698 & 0.5342 & 0.5151 & 0.4786 \\ 
XGBoost  & 0.6745 & 0.6132 & 0.5220 & 0.4762 \\
LightGBM & 0.7783 & 0.6462 & 0.5313 & 0.4913 \\ \bottomrule
\end{tabular}
\end{table}

%% file: main.bbl

\begin{thebibliography}{30}


\ifx \showCODEN    \undefined \def \showCODEN     #1{\unskip}     \fi
\ifx \showDOI      \undefined \def \showDOI       #1{#1}\fi
\ifx \showISBNx    \undefined \def \showISBNx     #1{\unskip}     \fi
\ifx \showISBNxiii \undefined \def \showISBNxiii  #1{\unskip}     \fi
\ifx \showISSN     \undefined \def \showISSN      #1{\unskip}     \fi
\ifx \showLCCN     \undefined \def \showLCCN      #1{\unskip}     \fi
\ifx \shownote     \undefined \def \shownote      #1{#1}          \fi
\ifx \showarticletitle \undefined \def \showarticletitle #1{#1}   \fi
\ifx \showURL      \undefined \def \showURL       {\relax}        \fi
\providecommand\bibfield[2]{#2}
\providecommand\bibinfo[2]{#2}
\providecommand\natexlab[1]{#1}
\providecommand\showeprint[2][]{arXiv:#2}

\bibitem[Chen et~al\mbox{.}(2021)]%
        {chen2021synthetic}
\bibfield{author}{\bibinfo{person}{Richard~J Chen}, \bibinfo{person}{Ming~Y
  Lu}, \bibinfo{person}{Tiffany~Y Chen}, \bibinfo{person}{Drew~FK Williamson},
  {and} \bibinfo{person}{Faisal Mahmood}.} \bibinfo{year}{2021}\natexlab{}.
\newblock \showarticletitle{Synthetic data in machine learning for medicine and
  healthcare}.
\newblock \bibinfo{journal}{\emph{Nature Biomedical Engineering}}
  \bibinfo{volume}{5}, \bibinfo{number}{6} (\bibinfo{year}{2021}),
  \bibinfo{pages}{493--497}.
\newblock


\bibitem[Chen and Guestrin(2016)]%
        {tianqi2016}
\bibfield{author}{\bibinfo{person}{Tianqi Chen} {and} \bibinfo{person}{Carlos
  Guestrin}.} \bibinfo{year}{2016}\natexlab{}.
\newblock \showarticletitle{{XGBoost: A scalable tree boosting system}}. In
  \bibinfo{booktitle}{\emph{KDD}}. \bibinfo{pages}{785--794}.
\newblock


\bibitem[DataCebo, Inc.(2023)]%
        {sdmetrics}
DataCebo, Inc. \bibinfo{year}{2023}\natexlab{}.
\newblock \bibinfo{booktitle}{\emph{Synthetic Data Metrics}}.
\newblock DataCebo, Inc.
\newblock
\urldef\tempurl%
\url{https://docs.sdv.dev/sdmetrics/}
\showURL{%
\tempurl}
\newblock
\shownote{Version 0.9.3}.


\bibitem[de~Masson~d'Autume et~al\mbox{.}(2019)]%
        {de2019training}
\bibfield{author}{\bibinfo{person}{Cyprien de Masson~d'Autume},
  \bibinfo{person}{Shakir Mohamed}, \bibinfo{person}{Mihaela Rosca}, {and}
  \bibinfo{person}{Jack Rae}.} \bibinfo{year}{2019}\natexlab{}.
\newblock \showarticletitle{Training language gans from scratch}.
\newblock \bibinfo{journal}{\emph{Advances in Neural Information Processing
  Systems}} (\bibinfo{year}{2019}).
\newblock


\bibitem[Dhariwal and Nichol(2021)]%
        {dhariwal2021diffusion}
\bibfield{author}{\bibinfo{person}{Prafulla Dhariwal} {and}
  \bibinfo{person}{Alex Nichol}.} \bibinfo{year}{2021}\natexlab{}.
\newblock \showarticletitle{Diffusion Models Beat {GANs} on Image Synthesis}.
\newblock \bibinfo{journal}{\emph{arXiv preprint arXiv:arXiv:2105.05233}}
  (\bibinfo{year}{2021}).
\newblock


\bibitem[Gianfrancesco et~al\mbox{.}(2018)]%
        {gianfrancesco2018potential}
\bibfield{author}{\bibinfo{person}{Milena~A Gianfrancesco},
  \bibinfo{person}{Suzanne Tamang}, \bibinfo{person}{Jinoos Yazdany}, {and}
  \bibinfo{person}{Gabriela Schmajuk}.} \bibinfo{year}{2018}\natexlab{}.
\newblock \showarticletitle{Potential biases in machine learning algorithms
  using electronic health record data}.
\newblock \bibinfo{journal}{\emph{JAMA Internal Medicine}}
  \bibinfo{volume}{178}, \bibinfo{number}{11} (\bibinfo{year}{2018}),
  \bibinfo{pages}{1544--1547}.
\newblock


\bibitem[Goodfellow et~al\mbox{.}(2020)]%
        {goodfellow2020generative}
\bibfield{author}{\bibinfo{person}{Ian Goodfellow}, \bibinfo{person}{Jean
  Pouget-Abadie}, \bibinfo{person}{Mehdi Mirza}, \bibinfo{person}{Bing Xu},
  \bibinfo{person}{David Warde-Farley}, \bibinfo{person}{Sherjil Ozair},
  \bibinfo{person}{Aaron Courville}, {and} \bibinfo{person}{Yoshua Bengio}.}
  \bibinfo{year}{2020}\natexlab{}.
\newblock \showarticletitle{Generative adversarial networks}.
\newblock \bibinfo{journal}{\emph{Commun. ACM}} \bibinfo{volume}{63},
  \bibinfo{number}{11} (\bibinfo{year}{2020}), \bibinfo{pages}{139--144}.
\newblock


\bibitem[Hwang(2022)]%
        {hwang2022}
\bibfield{author}{\bibinfo{person}{Chul-Hyun Hwang}.}
  \bibinfo{year}{2022}\natexlab{}.
\newblock \showarticletitle{Resolving CTGAN-based data imbalance for
  commercialization of public technology}.
\newblock \bibinfo{journal}{\emph{Journal of the Korea Institute of Information
  and Communication Engineering}} \bibinfo{volume}{26}, \bibinfo{number}{1}
  (\bibinfo{year}{2022}), \bibinfo{pages}{64--69}.
\newblock


\bibitem[Islam and Zhang(2020)]%
        {islam2020gan}
\bibfield{author}{\bibinfo{person}{Jyoti Islam} {and} \bibinfo{person}{Yanqing
  Zhang}.} \bibinfo{year}{2020}\natexlab{}.
\newblock \showarticletitle{GAN-based synthetic brain PET image generation}.
\newblock \bibinfo{journal}{\emph{Brain informatics}}  \bibinfo{volume}{7}
  (\bibinfo{year}{2020}), \bibinfo{pages}{1--12}.
\newblock


\bibitem[Jordon et~al\mbox{.}(2019)]%
        {jordon2019pate}
\bibfield{author}{\bibinfo{person}{James Jordon}, \bibinfo{person}{Jinsung
  Yoon}, {and} \bibinfo{person}{Mihaela Van Der~Schaar}.}
  \bibinfo{year}{2019}\natexlab{}.
\newblock \showarticletitle{PATE-GAN: Generating synthetic data with
  differential privacy guarantees}. In \bibinfo{booktitle}{\emph{International
  Conference on Learning Representations}}.
\newblock


\bibitem[Ke et~al\mbox{.}(2017)]%
        {LightGBM}
\bibfield{author}{\bibinfo{person}{Guolin Ke}, \bibinfo{person}{Qi Meng},
  \bibinfo{person}{Thomas Finley}, \bibinfo{person}{Taifeng Wang},
  \bibinfo{person}{Wei Chen}, \bibinfo{person}{Weidong Ma},
  \bibinfo{person}{Qiwei Ye}, {and} \bibinfo{person}{Tie-Yan Liu}.}
  \bibinfo{year}{2017}\natexlab{}.
\newblock \showarticletitle{{Light{GBM}: {A} Highly Efficient Gradient Boosting
  Decision Tree}}. In \bibinfo{booktitle}{\emph{Advances in Neural Information
  Processing Systems}}.
\newblock


\bibitem[Kim et~al\mbox{.}(2022)]%
        {gATE}
\bibfield{author}{\bibinfo{person}{Sundong Kim}, \bibinfo{person}{Tung duong
  Mai}, \bibinfo{person}{Sungwon Han}, \bibinfo{person}{Sungwon Park},
  \bibinfo{person}{Thi Nguyen}, \bibinfo{person}{Jaechan So},
  \bibinfo{person}{Karandeep Singh}, {and} \bibinfo{person}{Meeyoung Cha}.}
  \bibinfo{year}{2022}\natexlab{}.
\newblock \showarticletitle{{Active Learning for Human-in-the-loop Customs
  Inspection}}.
\newblock \bibinfo{journal}{\emph{IEEE Transactions on Knowledge and Data
  Engineering}} (\bibinfo{year}{2022}).
\newblock


\bibitem[Kim et~al\mbox{.}(2021)]%
        {kim2021KCS}
\bibfield{author}{\bibinfo{person}{Seongchan Kim}, \bibinfo{person}{Sa-Kwang
  Song}, \bibinfo{person}{Minhee Cho}, {and} \bibinfo{person}{Su-Hyun Shin}.}
  \bibinfo{year}{2021}\natexlab{}.
\newblock \showarticletitle{{Transaction Pattern Discrimination of Malicious
  Supply Chain using Tariff-Structured Big Data}}.
\newblock \bibinfo{journal}{\emph{The Journal of the Korea Contents
  Association}} (\bibinfo{year}{2021}).
\newblock


\bibitem[Kim et~al\mbox{.}(2020)]%
        {DATE}
\bibfield{author}{\bibinfo{person}{Sundong Kim}, \bibinfo{person}{Yu-Che Tsai},
  \bibinfo{person}{Karandeep Singh}, \bibinfo{person}{Yeonsoo Choi},
  \bibinfo{person}{Etim Ibok}, \bibinfo{person}{Cheng-Te Li}, {and}
  \bibinfo{person}{Meeyoung Cha}.} \bibinfo{year}{2020}\natexlab{}.
\newblock \showarticletitle{{DATE: Dual Attentive Tree-Aware Embedding for
  Customs Fraud Detection}}. In \bibinfo{booktitle}{\emph{KDD}}.
\newblock


\bibitem[Kong et~al\mbox{.}(2020)]%
        {kong2020hifi}
\bibfield{author}{\bibinfo{person}{Jungil Kong}, \bibinfo{person}{Jaehyeon
  Kim}, {and} \bibinfo{person}{Jaekyoung Bae}.}
  \bibinfo{year}{2020}\natexlab{}.
\newblock \showarticletitle{{Hifi-GAN}: Generative adversarial networks for
  efficient and high fidelity speech synthesis}. In
  \bibinfo{booktitle}{\emph{Advances in Neural Information Processing
  Systems}}.
\newblock


\bibitem[Lee et~al\mbox{.}(2021)]%
        {lee2021kaia}
\bibfield{author}{\bibinfo{person}{Eunji Lee}, \bibinfo{person}{Sundong Kim},
  \bibinfo{person}{Sihyun Kim}, \bibinfo{person}{Sungwon Park},
  \bibinfo{person}{Meeyoung Cha}, \bibinfo{person}{Soyeon Jung},
  \bibinfo{person}{Suyoung Yang}, \bibinfo{person}{Yeonsoo Choi},
  \bibinfo{person}{Sungdae Ji}, \bibinfo{person}{Minsoo Song}, {and}
  \bibinfo{person}{Heeja Kim}.} \bibinfo{year}{2021}\natexlab{}.
\newblock \showarticletitle{Classification of goods using text descriptions
  with sentences retrieval}. In \bibinfo{booktitle}{\emph{Korea Artificial
  Intelligence Conference (KAIA)}}.
\newblock


\bibitem[Mai et~al\mbox{.}(2021)]%
        {mai2021drift}
\bibfield{author}{\bibinfo{person}{Tung-Duong Mai}, \bibinfo{person}{Kien
  Hoang}, \bibinfo{person}{Aitolkyn Baigutanova}, \bibinfo{person}{Gaukhartas
  Alina}, {and} \bibinfo{person}{Sundong Kim}.}
  \bibinfo{year}{2021}\natexlab{}.
\newblock \showarticletitle{{Customs fraud detection in the presence of concept
  drift}}. In \bibinfo{booktitle}{\emph{ICDM IncrLearn Workshop}}.
\newblock


\bibitem[Mehrabi et~al\mbox{.}(2021)]%
        {mehrabi2021survey}
\bibfield{author}{\bibinfo{person}{Ninareh Mehrabi}, \bibinfo{person}{Fred
  Morstatter}, \bibinfo{person}{Nripsuta Saxena}, \bibinfo{person}{Kristina
  Lerman}, {and} \bibinfo{person}{Aram Galstyan}.}
  \bibinfo{year}{2021}\natexlab{}.
\newblock \showarticletitle{A survey on bias and fairness in machine learning}.
\newblock \bibinfo{journal}{\emph{ACM Computing Surveys (CSUR)}}
  \bibinfo{volume}{54}, \bibinfo{number}{6} (\bibinfo{year}{2021}),
  \bibinfo{pages}{1--35}.
\newblock


\bibitem[Mikuriya and Cantens(2020)]%
        {mikuriya2020wco}
\bibfield{author}{\bibinfo{person}{Kunio Mikuriya} {and}
  \bibinfo{person}{Thomas Cantens}.} \bibinfo{year}{2020}\natexlab{}.
\newblock \showarticletitle{{If Algorithms Dream of Customs, do Customs
  Officials Dream of Algorithms? A Manifesto for Data Mobilisation in
  Customs}}.
\newblock \bibinfo{journal}{\emph{World Customs Journal}} \bibinfo{volume}{14},
  \bibinfo{number}{2} (\bibinfo{year}{2020}).
\newblock


\bibitem[Nikolenko(2021)]%
        {nikolenko2021synthetic}
\bibfield{author}{\bibinfo{person}{Sergey~I Nikolenko}.}
  \bibinfo{year}{2021}\natexlab{}.
\newblock \bibinfo{booktitle}{\emph{Synthetic data for deep learning}}.
  Vol.~\bibinfo{volume}{174}.
\newblock \bibinfo{publisher}{Springer}.
\newblock


\bibitem[Pedregosa et~al\mbox{.}(2011)]%
        {scikit-learn}
\bibfield{author}{\bibinfo{person}{F. Pedregosa}, \bibinfo{person}{G.
  Varoquaux}, \bibinfo{person}{A. Gramfort}, \bibinfo{person}{V. Michel},
  \bibinfo{person}{B. Thirion}, \bibinfo{person}{O. Grisel},
  \bibinfo{person}{M. Blondel}, \bibinfo{person}{P. Prettenhofer},
  \bibinfo{person}{R. Weiss}, \bibinfo{person}{V. Dubourg}, \bibinfo{person}{J.
  Vanderplas}, \bibinfo{person}{A. Passos}, \bibinfo{person}{D. Cournapeau},
  \bibinfo{person}{M. Brucher}, \bibinfo{person}{M. Perrot}, {and}
  \bibinfo{person}{E. Duchesnay}.} \bibinfo{year}{2011}\natexlab{}.
\newblock \showarticletitle{Scikit-learn: Machine Learning in {P}ython}.
\newblock \bibinfo{journal}{\emph{Journal of Machine Learning Research}}
  \bibinfo{volume}{12} (\bibinfo{year}{2011}), \bibinfo{pages}{2825--2830}.
\newblock


\bibitem[Prokhorenkova et~al\mbox{.}(2018)]%
        {liudmila2018nips}
\bibfield{author}{\bibinfo{person}{Liudmila Prokhorenkova},
  \bibinfo{person}{Gleb Gusev}, \bibinfo{person}{Aleksandr Vorobev},
  \bibinfo{person}{Anna~Veronika Dorogush}, {and} \bibinfo{person}{Andrey
  Gulin}.} \bibinfo{year}{2018}\natexlab{}.
\newblock \showarticletitle{Cat{B}oost: unbiased boosting with categorical
  features support}. In \bibinfo{booktitle}{\emph{Advances in Neural
  Information Processing Systems}}. \bibinfo{pages}{6639--6649}.
\newblock


\bibitem[Qian et~al\mbox{.}(2023)]%
        {qian2023synthcity}
\bibfield{author}{\bibinfo{person}{Zhaozhi Qian},
  \bibinfo{person}{Bogdan-Constantin Cebere}, {and} \bibinfo{person}{Mihaela
  van~der Schaar}.} \bibinfo{year}{2023}\natexlab{}.
\newblock \showarticletitle{Synthcity: facilitating innovative use cases of
  synthetic data in different data modalities}.
\newblock \bibinfo{journal}{\emph{arXiv preprint arXiv:2301.07573}}
  (\bibinfo{year}{2023}).
\newblock


\bibitem[Singh et~al\mbox{.}(2023)]%
        {singh2023graphfc}
\bibfield{author}{\bibinfo{person}{Karandeep Singh}, \bibinfo{person}{Yu-Che
  Tsai}, \bibinfo{person}{Cheng-Te Li}, \bibinfo{person}{Meeyoug Cha}, {and}
  \bibinfo{person}{Shou-De Lin}.} \bibinfo{year}{2023}\natexlab{}.
\newblock \bibinfo{title}{GraphFC: Customs Fraud Detection with Label
  Scarcity}.
\newblock
\newblock
\showeprint[arxiv]{2305.11377}~[cs.LG]


\bibitem[Vanhoeyveld et~al\mbox{.}(2020)]%
        {vanhoeyveld2020belgian}
\bibfield{author}{\bibinfo{person}{Jellis Vanhoeyveld}, \bibinfo{person}{David
  Martens}, {and} \bibinfo{person}{Bruno Peeters}.}
  \bibinfo{year}{2020}\natexlab{}.
\newblock \showarticletitle{{Customs fraud detection: Assessing the value of
  behavioural and high-cardinality data under the imbalanced learning issue}}.
\newblock \bibinfo{journal}{\emph{Pattern Analysis and Applications}}
  \bibinfo{volume}{23} (\bibinfo{year}{2020}).
\newblock


\bibitem[Xu et~al\mbox{.}(2019)]%
        {CTGAN}
\bibfield{author}{\bibinfo{person}{Lei Xu}, \bibinfo{person}{Maria
  Skoularidou}, \bibinfo{person}{Alfredo Cuesta-Infante}, {and}
  \bibinfo{person}{Kalyan Veeramachaneni}.} \bibinfo{year}{2019}\natexlab{}.
\newblock \showarticletitle{{Modeling Tabular data using Conditional GAN}}. In
  \bibinfo{booktitle}{\emph{Advances in Neural Information Processing
  Systems}}.
\newblock


\bibitem[Xu and Veeramachaneni(2018)]%
        {xu2018synthesizing}
\bibfield{author}{\bibinfo{person}{Lei Xu} {and} \bibinfo{person}{Kalyan
  Veeramachaneni}.} \bibinfo{year}{2018}\natexlab{}.
\newblock \showarticletitle{Synthesizing Tabular Data using Generative
  Adversarial Networks}.
\newblock \bibinfo{journal}{\emph{arXiv preprint arXiv:1811.11264}}
  (\bibinfo{year}{2018}).
\newblock


\bibitem[Yoon et~al\mbox{.}(2020)]%
        {yoon2020anonymization}
\bibfield{author}{\bibinfo{person}{Jinsung Yoon}, \bibinfo{person}{Lydia~N
  Drumright}, {and} \bibinfo{person}{Mihaela Van Der~Schaar}.}
  \bibinfo{year}{2020}\natexlab{}.
\newblock \showarticletitle{Anonymization through data synthesis using
  generative adversarial networks (ads-gan)}.
\newblock \bibinfo{journal}{\emph{IEEE Journal of Biomedical and Health
  Informatics}} \bibinfo{volume}{24}, \bibinfo{number}{8}
  (\bibinfo{year}{2020}), \bibinfo{pages}{2378--2388}.
\newblock


\bibitem[Yoon et~al\mbox{.}(2019)]%
        {yoon2019time}
\bibfield{author}{\bibinfo{person}{Jinsung Yoon}, \bibinfo{person}{Daniel
  Jarrett}, {and} \bibinfo{person}{Mihaela Van~der Schaar}.}
  \bibinfo{year}{2019}\natexlab{}.
\newblock \showarticletitle{Time-series generative adversarial networks}. In
  \bibinfo{booktitle}{\emph{Advances in Neural Information Processing
  Systems}}.
\newblock


\bibitem[Yoon et~al\mbox{.}(2022)]%
        {yoon2022ehr}
\bibfield{author}{\bibinfo{person}{Jinsung Yoon}, \bibinfo{person}{Michel
  Mizrahi}, \bibinfo{person}{Nahid Ghalaty}, \bibinfo{person}{Thomas Jarvinen},
  \bibinfo{person}{Ashwin Ravi}, \bibinfo{person}{Peter Brune},
  \bibinfo{person}{Fanyu Kong}, \bibinfo{person}{Dave Anderson},
  \bibinfo{person}{George Lee}, \bibinfo{person}{Arie Meir}, {et~al\mbox{.}}}
  \bibinfo{year}{2022}\natexlab{}.
\newblock \showarticletitle{{EHR-Safe}: Generating High-Fidelity and
  Privacy-Preserving Synthetic Electronic Health Records}.
\newblock  (\bibinfo{year}{2022}).
\newblock


\end{thebibliography}
